\documentclass{article}

\usepackage[preprint]{neurips_2026}

\usepackage[utf8]{inputenc} 
\usepackage[T1]{fontenc}    
\usepackage{url}            
\usepackage{booktabs}       
\usepackage{amsfonts}       
\usepackage{nicefrac}       
\usepackage{microtype}      
\usepackage{xcolor}         
\usepackage{amsmath}
\usepackage{graphicx}

\usepackage[colorlinks=true,linkcolor=blue]{hyperref}

\usepackage{subcaption}
\usepackage{cleveref}
\usepackage{wrapfig}

\usepackage{algorithm}
\usepackage{algorithmicx}
\usepackage{algpseudocode}
\usepackage{multirow}
\usepackage{booktabs}

\newtheorem{rmk}{Remark}[section]

\title{Uncertainty-aware Feedback Loop Diagnosis: integrating reinforcement learning and conformal prediction for reliable wafer defect classification}

\author{%
  Aleksandr Dekhovich\thanks{Corresponding author. This work was done while the author was a postdoctoral researcher at TU Delft.} \\
  TU Delft\\
  \texttt{a.dekhovich@elsevier.com} \\
  \And
  Carlas Smith \\
  TU Delft\\
  \And
  Oleg Soloviev \\
  TU Delft \& Flexible Optical B.V.\\
}

\begin{document}

\maketitle

\begin{abstract}
  Semiconductor manufacturing is a complex, multi-step process that requires both high precision and low operating costs. Therefore, detecting defects in a high-volume setting is an important component of wafer map analysis for reliable production. However, most state-of-the-art deep learning approaches require a full scan of the surface, which hinders the development of high-throughput inspection systems. To address this problem, we propose Uncertainty-aware Feedback Loop Diagnosis (UFLD), a novel algorithm that employs reinforcement learning for adaptive measurement acquisition and provides a prediction uncertainty level after each step, stopping the process once sufficient information is obtained. Unlike existing methods that rely on ensemble models or auxiliary confidence estimation deep learning models, UFLD utilizes a more parameter-efficient conformal prediction framework to determine the optimal stopping time. The proposed UFLD reduces acquisition time by about $40\%$ and achieves $98.4\%$ inspection accuracy, without requiring an additional learned confidence-estimation network. We validated our approach on the MixedWM38 dataset, demonstrating its advantages over existing adaptive acquisition procedures and standard convolutional neural networks.
\end{abstract}

\section{Introduction}\label{sec1:intro}

The semiconductor industry’s evolution is historically driven by Moore’s Law, the empirical observation that the number of transistors in an integrated circuit doubles approximately every two years \cite{moore1965cramming}. Due to the increasing demand for chips, defect detection plays a crucial role in semiconductor manufacturing, ensuring the quality of chips produced \cite{azamfirei2023application}. However, this task is even more challenging in real-time high-volume manufacturing \cite{lee2018industrial}, which led to the development of automated defect inspection systems \cite{sandland1998automated, park2009automated}. In the past decade, artificial intelligence (AI), and specifically deep learning (DL), algorithms have achieved remarkable progress in a range of tasks, including industrial applications \cite{lee2018industrial, rai2021machine, alenizi2023artificial}. A significant portion of these works aims to develop vision-based defect detection systems to analyze the surface of interest by classifying \cite{fu2019deep, singh2023automated}, segmenting \cite{liu2022adaptive, dekhovich2025continual}, or detecting \cite{ma2024surface, zhu2023surface} possible defective regions.

\begin{figure}
    \centering
    \includegraphics[width=0.9\linewidth]{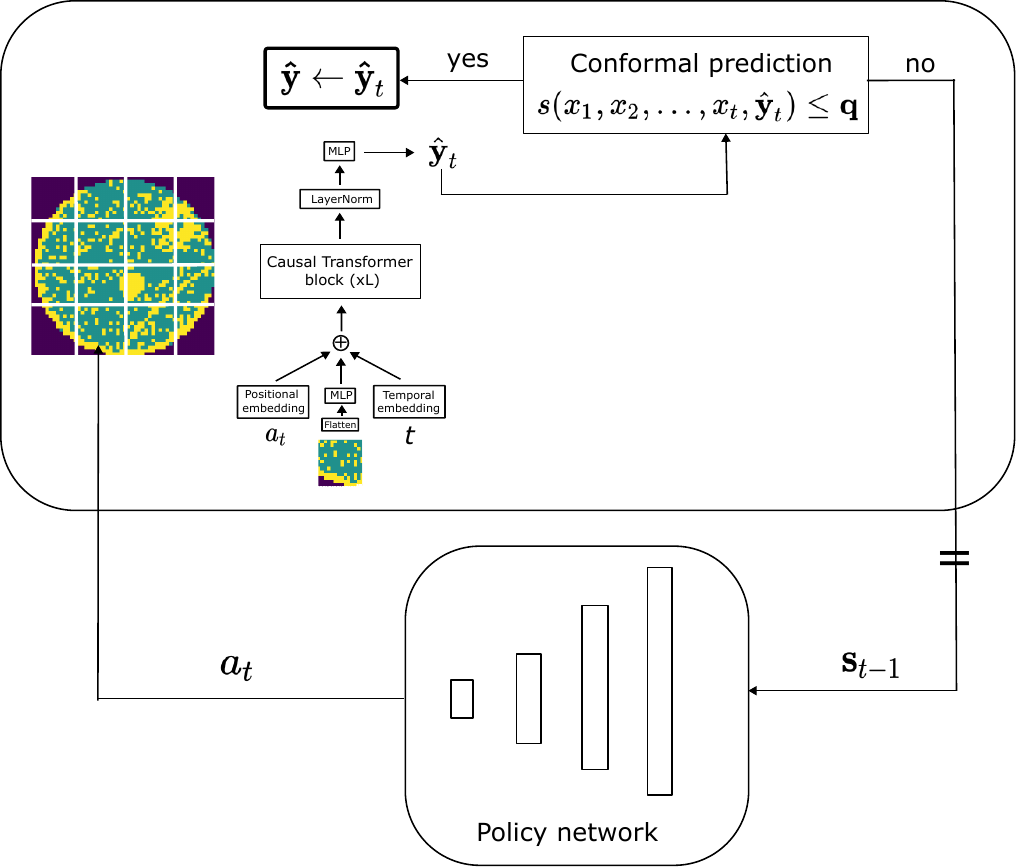}
    \caption{An overview of UFLD: the Causal Transformer predicts the defect type from the available patches; conformal prediction evaluates whether the predicted label is in the prediction set. If yes, the scanning process stops; otherwise, the policy network selects the next most informative patch, continuing the feedback loop.}
    \label{fig:ufld}
\end{figure}

An efficient inspection system should be accurate, reliable and fast at the same time. Therefore, despite the success of convolutional neural networks (CNNs) in classification problems, CNN-based solutions typically require a full scan of the wafer. In modern high-volume manufacturing, collecting high-resolution data for every die is expensive and time-consuming. Consequently, there is a need for alternative approaches employing adaptive data acquisition (ADA) strategies that can maintain high classification accuracy while minimizing the number of scanned regions. 

To address this, we propose Uncertainty-aware Feedback Loop Diagnosis (UFLD) that uses a reinforcement learning (RL) strategy for ADA and a conformal prediction-based framework \cite{vovk2005algorithmic} to evaluate the trustworthiness of the prediction after every iteration of the acquisition process to mitigate the limitations of the previous works \cite{dekhovich2025sequential, dekhovich2025supconwi}. \Cref{fig:ufld} illustrates the algorithm: the Causal Transformer plays the role of the classifier to predict the defect pattern from the available observations, while the RL-based policy network adaptively selects the next region for scanning. After each step, a simple threshold-based rule determines whether the acquisition process can be stopped. The main contributions of our work can be summarized as follows:
\begin{itemize}
    \item We developed a feedback loop algorithm to increase the throughput of inspection tools, while maintaining the performance level of CNN models that require full scanning. The code implementation of the work is publicly available on GitHub: \url{https://github.com/adekhovich/UFLD}.
    \item In contrast to previous works on sequential wafer inspection where RNN-based classifiers were used, we utilize a causal transformer-based model that has better predictive capabilities. A single transformer model achieves performance comparable to an ensemble of RNNs, while requiring fewer training parameters.
    \item We introduce a novel stopping rule for the adaptive data acquisition process based on the conformal prediction framework. Our approach reduces algorithmic complexity compared to previous works by replacing an auxiliary confidence-estimator network with a simple threshold-based rule without sacrificing prediction quality. 
\end{itemize}

The rest of the article is organized as follows: in Section \ref{sec2:related_works}, we provide the necessary background for the work, Section \ref{sec3:method} describes the proposed feedback loop algorithm, while Section \ref{sec4:experiments} shows the results and compares with other methods, and Section \ref{sec5:conclusion} concludes the article.

\section{Related work}\label{sec2:related_works}

\paragraph{Wafer defect recognition using deep learning}
Deep learning (DL) models are widely used in wafer failure pattern detection \cite{nakazawa2018wafer, yu2023wafer, yuan2025position}. Convolutional neural networks (CNNs) have been trained under different scenarios, namely zero- and few-shot learning \cite{kim2024generalized, geng2021wafer} to meet the real-world data scarcity constraints, as well as transfer learning \cite{yu2021wafer, chou2025adaptive} and supervised contrastive learning \cite{bae2023supervised} for better generalization. In addition, some works consider unsupervised learning tasks for wafer map analysis with CNNs \cite{kim2023advances}. For instance, the active cluster framework \cite{shim2021active} has been proposed for annotating unlabeled wafer maps. Furthermore, Nakazawa \& Kulkarni \cite{nakazawa2018wafer} developed a CNN-based encoder–decoder model for abnormal defect pattern detection and segmentation of previously unseen real wafer maps.

Transformer-based architectures have been successfully used in defect classification tasks \cite{fan2024new, jain2026sdcvit}. By leveraging self-attention layers, ViTs \cite{dosovitskiyimage} often generalize better than CNNs; however, they may require more training data for convergence \cite{lafrance2025evaluating}. Therefore, fine-tuning of the pretrained ViT is a better strategy than training a model from scratch \cite{mohammad2025semiconductor, chen2025wafer}. Recently, Liao \emph{et al}. proposed WaferDSR-RAG \cite{liao2026wafer}, which uses retrieval-augmented generation (RAG) \cite{lewis2020retrieval}. 
In WaferDSR-RAG, this technique is used for an LLM with a specialized semiconductor knowledge base, linking visual defect patterns to historical engineering logs. 

In the discussed works, both CNNs and Transformer-based architectures require a full wafer map and do not process partial observations from the surface, which makes them impractical for real-time high-volume manufacturing. Consequently, we further discuss approaches that predict a label by adaptively selecting the most informative regions, which significantly reduces overall scanning time.

\paragraph{Adaptive data acquisition}
Adaptive data acquisition is a widely used technique to reduce the cost of measurement collection in high-cost processes. In scanning microscopy, Ye \emph{et al.}~\cite{ye2025learned} introduced a closed-loop denoising system with pixel-wise uncertainty estimation based on conformal prediction, where only the most uncertain regions of an image are adaptively selected for rescanning. Holman \emph{et al.}~\cite{holman2020autonomous} developed a grid-less autonomous adaptive sampling method for scanning hyperspectral imaging. By utilizing hybrid sequential design strategies for surrogate model construction, their system identifies regions of steeper physico-chemical gradients and increases sampling density in those areas in real-time. This approach significantly reduces acquisition time compared to traditional uniform grid sampling while maintaining high fidelity in complex biological samples.

Several works employ reinforcement learning (RL) \cite{sutton1998introduction} for adaptive data acquisition \cite{betterton2020reinforcement, yen2024adaptive, dekhovich2025sequential}, modeling the problem as a Markov Decision Process (MDP) defined by the tuple $(\mathcal{S}, \mathcal{A}, P, R, \mu, \gamma)$. In this framework, $\mathcal{S}$ represents the state space, $\mathcal{A}$ the action space, $P(s_{t+1}|s_t, a_t)$ the transition distribution, $R(s_t, a_t)$ the reward function, $\mu(s_0)$ the initial state distribution, and $\gamma \in (0, 1]$ the discount factor. The agent’s objective is to learn an optimal policy $\pi: \mathcal{S} \times \mathcal{A} \to [0, 1]$ that maximizes the expected cumulative return, given by $\mathbb{E}_{\pi}\left[\sum_{t} \gamma^{t} R(s_t, a_t)\right]$, where actions and next states are sampled according to $a_t \sim \pi(\cdot | s_t)$ and $s_{t+1} \sim P(\cdot | s_t, a_t)$, respectively.

For instance, Bakker \emph{et al}. \cite{bakker2020experimental} utilize a policy gradient RL method for adaptive sampling in $k$-space for MRI reconstruction, while Yen \emph{et al}. \cite{yen2024adaptive} developed Adaptive Sampling for MR (ASMR) for image-based disease detection. However, these methods predefine the number of measurements that are adaptively selected. The recently proposed SupConWI-RL \cite{dekhovich2025supconwi} adaptively selects the next patch and stops the process when an auxiliary confidence estimator model signals that the collected information is sufficient for a reliable prediction. As a result, the number of scanned patches is not predefined and depends on how difficult the failure pattern is to classify.

\paragraph{Conformal prediction}
Conformal prediction (CP) is a data-based, distribution-free technique for constructing prediction sets, using a calibration set $\mathcal{D}_{\text{calib}} = \{(x_i, y_i)\}_{i=1}^{n_{\text{calib}}}$ that is different from the train or test set \cite{shafer2008tutorial}. The key assumption for the coverage guarantee is the exchangeability property. In addition, it is worth noting that the i.i.d. assumption is often used since exchangeability follows from i.i.d., while the opposite is not true. Vanilla CP consists of the following steps: 

\begin{enumerate}
    \item select a non-conformity scoring function $s: X \times Y \to [0, 1]$;
    \item calculate non-conformity scores $s_1, s_2, \ldots, s_{n_{\text{calib}}}$, where $s_i = s(x_i, y_i), \ (x_i, y_i) \in \mathcal{D}_{\text{calib}}$;
    \item calculate the empirical quantile $\hat{q}_{\alpha} = \frac{\lceil (n_{\text{calib}} + 1)(1 - \alpha) \rceil}{n_{\text{calib}}}$ of the scores $s_1, s_2, \dots, s_{n_{\text{calib}}}$, where $1-\alpha$ is the user-defined coverage level;
    \item construct the prediction set for a test point $x_{\text{test}}$ with a miscoverage level $\alpha$: $C_{\alpha}(x_{\text{test}}) = \{y: \ s(x_{\text{test}}, y) \le \hat{q}_{\alpha}\}$.
\end{enumerate}

Vovk \emph{et al.} \cite{vovk2005algorithmic} provide theoretical guarantees for the prediction sets depending on the size of the calibration set $n_{\text{calib}}$:
$
\mathbb{P}\big(y \in C_{\alpha}(x_{\text{test}})\big) \ge 1-\alpha,
$
for the true label $y$ of the test point $x_{\text{test}}$. Adaptive Prediction Sets (APS) \cite{romano2020classification, angelopoulos2020uncertainty} and Conformal Risk Control (CRC) \cite{angelopoulos2022conformal} have been proposed as extensions of the vanilla CP. Like vanilla CP, APS guarantees marginal coverage, while its adaptive conformity score allows the prediction-set size to vary with the difficulty of each input. Furthermore, CRC generalizes vanilla CP to the scenario where the output is high-dimensional by controlling an average loss, e.g., an image segmentation task with a binary mask.

\section{Proposed method: Uncertainty-aware feedback loop diagnosis}\label{sec3:method}

Our Uncertainty-aware Feedback Loop Diagnosis (UFLD) framework consists of three core components: a Transformer-based classifier, an RL-based adaptive patch selector, and a conformal prediction-based stopping rule. The goal of UFLD is to optimize the input sequence to minimize the number of measurements. To simulate real-world conditions, we split an input image, a wafer map, into patches that can be treated as dies. Formally, we denote the input as $\mathbf{X} = (\mathbf{x}_1, \mathbf{x}_2, \ldots, \mathbf{x}_T)$, where $\mathbf{x}_i$ is a patch. In our algorithm, the RL-based selector iteratively selects the next patch based on prior observations, while the Transformer-based classifier predicts the defect type after each acquisition. The loop continues until the conformal prediction-based stopping rule determines that the classification reaches a reliability threshold.


\begin{figure}
    \centering
    \includegraphics[width=0.9\linewidth]{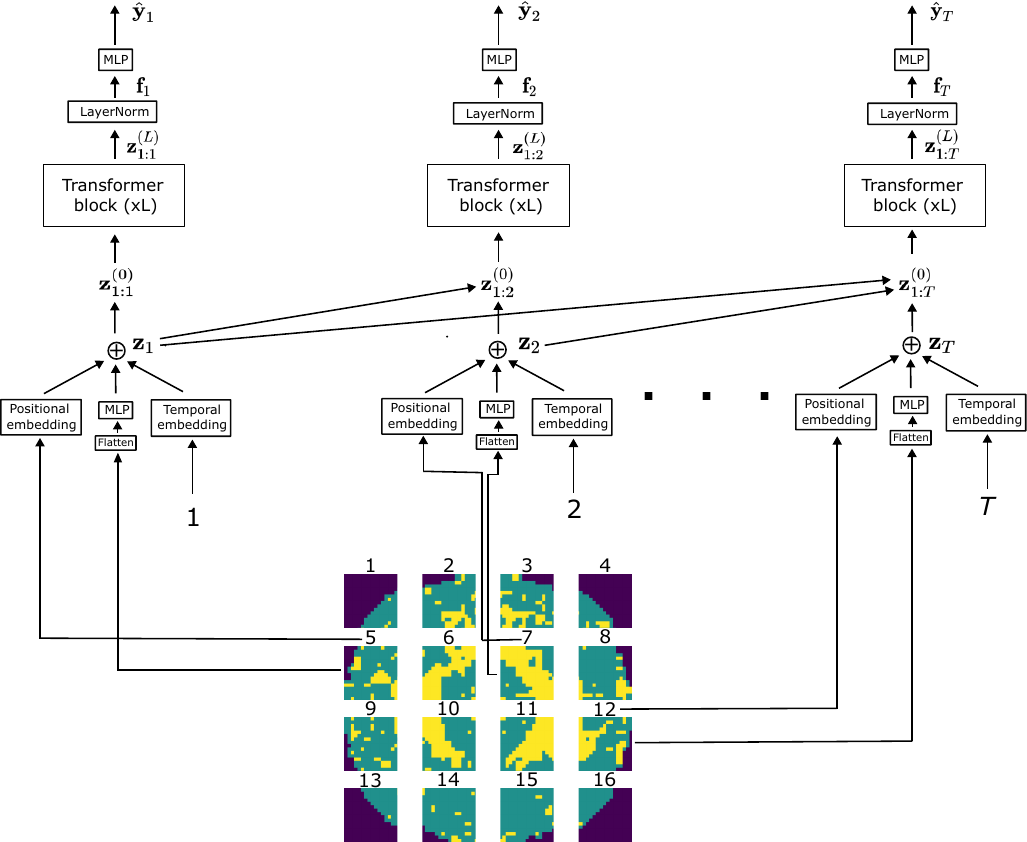}
    \caption{An illustration of the transformer-based classifier: at every time step $t$ it aggregates $t$ embedded patches along with their positional and temporal embedding, followed by $L$ Transformer blocks. Finally, the layer normalization and MLP are applied to obtain the output.}
    \label{fig:ufld-transformer}
\end{figure}

\paragraph{Transformer-based classifier} First, we pretrain our classifier on randomly sampled sequences of patches $\mathbf{x}_{i_1}, \mathbf{x}_{i_2}, \ldots, \mathbf{x}_{i_T}$, so the Transformer model can provide accurate predictions across different acquisition trajectories that the RL agent explores during optimization. To achieve this sequence robustness, a standard causal Transformer is augmented with a learnable sequence-step embedding $\mathbf{E}_{t}$. This embedding encodes the temporal position of a patch within the acquisition sequence, supplementing spatial positional embeddings $\mathbf{E}_{i_t}$ of the patch on the wafer map. \Cref{fig:ufld-transformer} represents the architecture of the classifier. Overall, the formal computations within the model to compute output $\hat{\mathbf{y}}_{t}$ for $1 \le t \le T$ are defined as follows:

\begin{equation}
    \begin{aligned}
        \mathbf{z}_{s} &=  \texttt{MLP}(\texttt{Flatten}(\mathbf{x}_{i_s})) + \mathbf{E}_{i_s} + \mathbf{E}_{s}, \quad 1 \le s \le t  \\
        \mathbf{z}_{1:t}^{(0)} &:= [\mathbf{z}_{1} \ \mathbf{z}_{2} \ldots \mathbf{z}_{t}] \\
        \mathbf{z}_{1:t}^{(l)} &= \texttt{TransformerBlock}\bigg(\mathbf{z}_{1:t}^{(l-1)}\bigg), \quad l=1, 2, \ldots, L ,        \\
        \mathbf{f}_t &= \texttt{LayerNorm}\bigg(\mathbf{z}_{1:t}^{(L)}\bigg) \\
        \hat{\mathbf{o}}_{t} &= \texttt{MLP}\big(\mathbf{f}_t\big). \\
     \end{aligned}
\end{equation}

Since each wafer map can contain up to four defects in the MixedWM38 dataset, we are solving a multi-label classification problem. To address this, we apply multi-hot encoding to the labels, resulting in a target vector $\mathbf{y} \in \{0, 1\}^C$, where $C$ denotes the total number of possible defect classes. Therefore, the output of the classifier is a probability tensor $\mathbf{p}_{t}$ that is calculated by applying the \texttt{Sigmoid} function to each component of the logit tensor $\hat{\mathbf{o}}_{t}$: $\mathbf{p}_{t} = \texttt{Sigmoid}(\hat{\mathbf{o}}_{t})$.

Then, we obtain the prediction $\hat{\mathbf{y}}_{t}$ by thresholding the probability tensor $\mathbf{p}_{t} = (p_{t, 1}, p_{t, 2}, \ldots, p_{t, C})$:
$
\hat{\mathbf{y}}_{t} = \big(\mathbf{1}\{p_{t, 1} \ge 0.5\}, \mathbf{1}\{p_{t, 2} \ge 0.5\}, \ldots, \mathbf{1}\{p_{t, C} \ge 0.5\}\big).
$

To optimize the parameters of the Transformer, we use standard binary cross-entropy loss for each type of defect. Several works have shown the effectiveness of the supervised contrastive learning loss term for the classification task \cite{khosla2020supervised}, including sequential wafer map defect recognition \cite{bae2023supervised, dekhovich2025continual}. Therefore, we include it in the total loss that we optimize:
\begin{multline*}
\mathcal{L}(\theta_{\text{classifier}}) = \frac{1}{|\mathcal{B}_{\text{aug}}|}\frac{1}{T}\sum_{(\mathbf{X}, \mathbf{y}) \in \mathcal{B}_{\text{aug}}} \sum_{t=1}^T \bigg(\text{BCELoss}(\theta_{\text{classifier}}; \mathbf{p}_{t}, \mathbf{y}) + \\\lambda \cdot \text{SCLoss}(\theta_{\text{classifier}}; \{\mathbf{f}_t\}_{i=1}^{|\mathcal{B}|}, \{\tilde{\mathbf{f}}_t\}_{i=1}^{|\mathcal{B}|}, \mathbf{y}) \bigg),
\end{multline*}
where batch $\mathcal{B}_{\text{aug}} = \mathcal{T}_1(\mathcal{B}) \cup \mathcal{T}_2(\mathcal{B})$ is obtained by applying two data augmentations $\mathcal{T}_1$ and $\mathcal{T}_2$ to the original training batch $\mathcal{B}$, $\{\mathbf{f}_t\}_{i=1}^{|\mathcal{B}|}$ and $\{\tilde{\mathbf{f}}_t\}_{i=1}^{|\mathcal{B}|}$ are the last feature representations of sequences up to time $t$ of $\mathcal{T}_1(\mathcal{B})$ and $ \mathcal{T}_2(\mathcal{B})$ respectively.

\paragraph{RL-based patch selection} Once the classifier is trained, our goal is to optimize the acquisition sequence to minimize the number of measurements (scanned patches) required for each wafer map. Therefore, we train a policy network $\pi_{\theta_{\text{RL}}}$ that selects the next patch from the set of available candidates. During training, the next action (patch) $a_t$ is sampled from a categorical distribution with parameters $\mathbf{\pi}_{t+1} =  \pi_{\theta_{\text{RL}}}(\mathbf{s}_t) \odot \mathbf{m}_t$: $a_{t+1} \sim \text{Categorical}(\mathbf{\pi}_{t+1})$,
where $\mathbf{m}_t \in \{0, 1\}^T$ is a binary mask representing the remaining unselected patches and $\mathbf{s}_t = (\mathbf{f}_t, \mathbf{m}_t)$. At initial step $t=0$, $\mathbf{m}_0 = (1, 1, \ldots, 1)$ and $\mathbf{f}_0  = (0, 0, \ldots, 0)$. During inference, the agent follows a deterministic policy by selecting the available action with the maximum probability: $a_t = \text{argmax} \ {\mathbf{\pi}_t}.$
The reward function is designed as follows: 
$
    r_t = 
    \begin{cases}
       1 - c_{\text{RL}}, \quad \text{if} \ \hat{\mathbf{y}}_t = \mathbf{y}\\
       - c_{\text{RL}}, \quad \text{otherwise},
    \end{cases}
$
where $c_{\text{RL}}$ is a penalty term for every step. We should note that our reward function is different from the one in \cite{dekhovich2025supconwi} because ours only depends on the correctness of the prediction, and not the score of some auxiliary confidence estimator, which may be unreliable or poorly calibrated.

\paragraph{Conformal prediction as stopping rule} The main challenge in the proposed acquisition process is determining the stopping time $t$ where the classification model is sufficiently certain to use the prediction $\hat{\mathbf{y}}_t$. Stopping too early may result in inaccurate predictions, while delaying the stopping point will increase the data collection costs. The conformal prediction framework provides the opportunity to control the coverage rate $1-\alpha$. However, the application of conformal prediction to sequential decision-making is complicated by the potential violation of the exchangeability property, which is the fundamental assumption for the calibration and test datasets. In our case, the predictions $\hat{\mathbf{y}}_t$ are computed from partial observations $\mathbf{x}_{1:t}$. We ensure that the calibration set $\mathcal{D}_{\text{calib}}$ is processed in the same manner as the test set $\mathcal{D}_{\text{test}}$. Therefore, we should calculate the non-conformity scores and, as a result, the threshold $\hat{q}_{\alpha}$ on the calibration set using trajectories generated by the same policy across the same time steps as the test set.

In our work, we focus on point accuracy, where each component of the predicted label is correct. Therefore, our objective is the per-class certainty in the classifier output $\hat{\mathbf{y}}_{t, i}, \ i=1, 2, \ldots, C$ at time step $t$. For a given calibration point $(\mathbf{x}^{i}, \mathbf{y}^{i}) \in \mathcal{D}_{\text{calib}}$, we define non-conformity scores at time step $t$ as follows:
\begin{equation}\label{eq:nonconf_scores}
S^{i}_{t,j}(\mathbf{x}^{i}, \mathbf{y}^{i}) = \begin{cases}
    1 - p^{i}_{t,j}, \quad \text{if} \ y^{i}_{j} = 1\\
    p^{i}_{t,j}, \quad \ \ \quad \text{if} \ y^{i}_{j} = 0,
\end{cases}
\end{equation}
where $p^i_{t,j}$ is the $j$-th component of the classifier's output probability distribution $\mathbf{p}_{t}$, and $\mathbf{y}^{i} = (y^{i}_{1}, y^{i}_{2}, \ldots, y^{i}_{C})$.

Our approach relies on two sets of label-wise quantiles $\hat{\mathbf{q}}_y = (\hat{q}_{y,1}, \hat{q}_{y,2}, \dots, \hat{q}_{y,C}), \ y\in \{0, 1\}$ for positive and negative classes. These quantiles play the role of thresholds, and we further call them "negative" and "positive" thresholds. These thresholds are computed through the calibration procedure. During inference, for a new test input $\mathbf{x}$, we calculate non-conformity scores 
\begin{equation}\label{eq:nonconf_test}
        \begin{aligned}[b]
        S_{0, t, j}(\mathbf{x}_{1:t}) &= p_{t,j}, \\
        S_{1, t, j}(\mathbf{x}_{1:t}) &= 1-p_{t,j}  
        \end{aligned}
\end{equation}
after every acquisition step $t$. We simultaneously track these values for each label $j=1, 2, \ldots, C$, and our goal is to stop when the prediction set for each label has cardinality $1$. Let us denote $E_{t,j} := \{S_{0, t, j} \le \hat{q}_{0,j} \land S_{1, t, j} > \hat{q}_{1,j}\} \lor \{S_{0, t, j} > \hat{q}_{0,j} \land S_{1, t, j} \le \hat{q}_{1,j}\}$ for each label $j=1,2,\ldots,C$. Therefore, our requirement of the size of the prediction set for label $j$ is equivalent to $E_{t,j} = 1$. We need to stop the acquisition process when we are sure about all the predicted labels, which means that $E_t:= E_{t,1} \land E_{t,2} \land \ldots \land E_{t,C} = 1$, and, therefore, every event $E_{t,j}$ is true. This means that the non-conformity score of every label is lower than the corresponding component of precisely one of the thresholds $\hat{\mathbf{q}}_0, \hat{\mathbf{q}}_1$. As a result, we can formulate our RL acquisition process with the conformal prediction stopping rule in \Cref{alg:alg_final}.

\begin{algorithm}[ht]
    \begin{algorithmic}[1]
    \Function{Predict}{transformer classifier $\mathcal{F}$, acquisition policy $\pi$, test point $\mathbf{x}$, quantiles $\hat{\mathbf{q}}_{0}, \hat{\mathbf{q}}_{1}$}

    \State{$\hat{\mathbf{y}} \gets \emptyset$}
    \State{$\mathbf{m}_0 \gets (1, 1, \ldots, 1)$}      
    \State{$\mathbf{f}_0 \gets (0, 0, \ldots, 0)$}
    \For{$t=0, 1, \ldots, T-1$} 
        \State{Initialize current state: $\mathbf{s}_t \gets [\mathbf{f}_t, \mathbf{m}_t]$} 
        \State{Predict next patch location: $a_{t+1} \gets \pi(\mathbf{s}_t)$}    
        \State{$\mathbf{m}_{t+1} \gets \mathbf{m}_t$}
        \State{$\mathbf{m}_{t+1}[a_{t+1}] \gets 0$}
        \State{$\mathbf{x}_{1:{t+1}} \gets [x_{a_{1}}, x_{a_{2}}, \ldots, x_{a_{t+1}}]$}
        \State{$\hat{\mathbf{y}}_{t+1}, \mathbf{f}_{t+1} \gets \mathcal{F}(\mathbf{x}_{1:{t+1}})$}

        \State{Compute $\mathbf{S}_{0, t+1}(\mathbf{x}_{1:{t+1}})$ and $\mathbf{S}_{1, t+1}(\mathbf{x}_{1:{t+1}})$ using (\ref{eq:nonconf_test})}
        \If{$E(\mathbf{S}_{0, t+1}, \mathbf{S}_{1, t+1}, \hat{\mathbf{q}}_{0}, \hat{\mathbf{q}}_{1}) = 1$}
            \State{$\hat{\mathbf{y}} \gets \hat{\mathbf{y}}_{t+1}$}
            \State{break loop}
        \EndIf
        
    \EndFor
    \If{$\hat{\mathbf{y}} = \emptyset$}
        \State{$\hat{\mathbf{y}} \gets \mathcal{F}(\mathbf{x}_{1:{T}})$}
    \EndIf   
    
    \State{\Return $\hat{\mathbf{y}}$}
    \EndFunction
    \end{algorithmic}
    \caption{Inference with adaptive data acquisition and conformal prediction}
    \label{alg:alg_final}
\end{algorithm}

Due to the sequential nature of our problem and differences in stopping times for each of the wafers, the quantiles $\hat{\mathbf{q}}_{0}, \hat{\mathbf{q}}_{1}$ have to be computed from the non-conformity scores obtained from different time steps to better align the calibration setting with inference. However, at first, we do not have a rule to stop the acquisition process and collect the non-conformity scores for calibration. Thus, our calibration process consists of two parts: first, we compute the preliminary quantiles $\hat{\mathbf{q}}^A_{0}, \hat{\mathbf{q}}^A_{1}$ using the dataset $\mathcal{D}_A \subset \mathcal{D}_{\text{calib}}$ and by collecting the non-conformity scores $\mathbf{S}^i_{T}$ at the final step $T$. Second, we recalibrate these quantiles using them as a stopping criterion for the learned acquisition policy $\pi$ on dataset $\mathcal{D}_B = \mathcal{D}_{\text{calib}} \setminus \mathcal{D}_A$. For each wafer in $\mathcal{D}_B$, non-conformity scores are then collected at the induced stopping time and used to compute the refined thresholds $\hat{\mathbf{q}}_{0}, \hat{\mathbf{q}}_{1}$. \Cref{alg:calibrate} combines these stages into the full calibration algorithm. In this way, the calibration process reproduces the inference setting, where predictions may be produced at input-dependent stopping times. \Cref{alg:calibrate_A} illustrates the first stage of the calibration process, where we collect the non-conformity scores at time $T$. In \Cref{alg:calibrate_B}, we use $\hat{\mathbf{q}}^A_{0} $ and $ \hat{\mathbf{q}}^A_{1}$ to construct a stopping rule and to collect the non-conformity scores at the end of the scanning process, which can be at an arbitrary time step.

\begin{algorithm}[ht]
    \caption{Calibrate at time step $T$}
    \label{alg:calibrate_A}
    \begin{algorithmic}[1]
        \Function{Calibrate\_A}{transformer classifier $\mathcal{F}$, calibration dataset $\mathcal{D}_{A}=\{(\mathbf{x}^i, \mathbf{y}^i)\}_{i=1}^{n_A}$, miscoverage rate $\alpha$}
            \State Prediction at final step $T$: $\hat{\mathbf{y}}^i = \mathcal{F}(\mathbf{x}^i) \quad i=1,2,\ldots,n_A$
            \State Compute non-conformity scores $\mathbf{S}^{i}_y = (S^i_{y,1}, S^i_{y,2}, \ldots, S^i_{y,C})$, $y=0,1$, using \Cref{eq:nonconf_scores}
            \State Compute per-class quantiles $\hat{\mathbf{q}}^A_{y} = (\hat{q}^A_{y,1}, \hat{q}^A_{y,2}, \ldots, \hat{q}^A_{y,C})$, \quad where:
            \Statex \hspace{\algorithmicindent} $\hat{q}^A_{y,j} = \text{Quantile}\bigg(\{S^i_{y,j}: y_j^i=y\}_{i=1}^{n_{\text{A}_{y, j}}}; \frac{\lceil (n_{\text{A}_{y, j}} + 1)(1 - \alpha) \rceil}{n_{\text{A}_{y, j}}}\bigg), \ \ n_{\text{A}_{y,j}}=\left|\left\{i:y_j^i=y\right\}\right|$
            \State \Return $\hat{\mathbf{q}}^A_{0}, \hat{\mathbf{q}}^A_{1}$
        \EndFunction
    \end{algorithmic}
\end{algorithm}

\begin{algorithm}[ht]
    \begin{algorithmic}[1]
    \Function{Calibrate\_B}{transformer classifier $\mathcal{F}$, acquisition policy $\pi$, calibration dataset $\mathcal{D}_{B}=\{(\mathbf{x}^i, \mathbf{y}^i)\}_{i=1}^{n_B}$, preliminary quantiles $\hat{\mathbf{q}}^A_{0}, \hat{\mathbf{q}}^A_{1}$, miscoverage rate $\alpha$}
    \State{/* Set of "negative" non-conformity scores */}
    \State{$S_0[j] \gets \emptyset, \ j=1,2, \ldots, C.$}
    \State{/* Set of "positive" non-conformity scores */}
    \State{$S_1[j] \gets \emptyset, \ j=1,2, \ldots, C.$}

    \For{$i=1, 2, \ldots, n_B$}
        \State{$\mathbf{m}_0 \gets (1, 1, \ldots, 1)$}      
        \State{$\mathbf{f}_0 \gets (0, 0, \ldots, 0)$}
        \For{$t=0, 1, \ldots, T-1$}
            \State{Initialize current state: $\mathbf{s}_t \gets [\mathbf{f}_t, \mathbf{m}_t]$} 
            \State{Predict next patch location: $a_{t+1} \gets \pi(\mathbf{s}_t)$}    
            \State{$\mathbf{m}_{t+1} \gets \mathbf{m}_t$}
            \State{$\mathbf{m}_{t+1}[a_{t+1}] \gets 0$}
            \State{$\mathbf{x}^i_{1:{t+1}} \gets [x^i_{a_{1}}, x^i_{a_{2}}, \ldots, x^i_{a_{t+1}}]$}
            \State{$\hat{\mathbf{y}}^i_{t+1}, \mathbf{f}_{t+1} \gets \mathcal{F}(\mathbf{x}^i_{1:{t+1}})$}
            \State{Compute $\mathbf{S}^i_{0, t+1}(\mathbf{x}^i_{1:{t+1}})$ and $\mathbf{S}^i_{1, t+1}(\mathbf{x}^i_{1:{t+1}})$ using (\ref{eq:nonconf_test})}
        \If{$E(\mathbf{S}^i_{0, t+1}, \mathbf{S}^i_{1, t+1}, \hat{\mathbf{q}}^A_{0}, \hat{\mathbf{q}}^A_{1}) = 1$ \textbf{or} $t = T-1$}
            \State{/* Collect non-conformity scores at time $t+1$ */}
            \For{$j=1, 2, \ldots, C$}
                \If{$y^i_j = 0$}
                    \State{$S_0[j] \gets S_0[j]\cup \{S^i_{0, t+1, j}\}$}
                \Else
                    \State{$S_1[j] \gets S_1[j]\cup \{S^i_{1, t+1, j}\}$}
                \EndIf
            \EndFor
            \State{\textbf{break}}
        \EndIf
        \EndFor
    \EndFor

    \State{/* Compute refined per-class quantiles $\hat{\mathbf{q}}^B_{0}, \hat{\mathbf{q}}^B_{1}$ */}
    \State{$\hat{q}^B_{y,j} = \text{Quantile}\bigg(S_y[j];\frac{\lceil (n_{\text{B}_{y,j}} + 1)(1 - \alpha) \rceil}{n_{\text{B}_{y,j}}}\bigg), \ y=0,1, \ j=1, 2, \ldots, C.$}
    
    \State{\Return $\hat{\mathbf{q}}^B_{0}, \hat{\mathbf{q}}^B_{1}$}
    \EndFunction
    \end{algorithmic}
    \caption{Recalibrate using policy $\pi$}
    \label{alg:calibrate_B}
\end{algorithm}

\begin{algorithm}[ht]
    \begin{algorithmic}[1]
    \Function{Calibrate}{transformer classifier $\mathcal{F}$, acquisition policy $\pi$, calibration dataset $\mathcal{D}_{\text{calib}}=\{(\mathbf{x}^i, \mathbf{y}^i)\}_{i=1}^{n_{\text{calib}}}$, miscoverage rate $\alpha$}

    \State{Split $\mathcal{D}_{\text{calib}}$ into $\mathcal{D}_A$ and $\mathcal{D}_B$}
    \State{/* Compute quantiles using final time step T */}
    \State{$\hat{\mathbf{q}}^A_{0}, \hat{\mathbf{q}}^A_{1} \gets \text{Calibrate\_A}(\mathcal{F}, \mathcal{D}_A, \alpha)$}
    \State{/* Recalibrate quantiles using policy $\pi$ */}
    \State{$\hat{\mathbf{q}}_{0}, \hat{\mathbf{q}}_{1} \gets \text{Calibrate\_B}(\mathcal{F}, \pi, \mathcal{D}_B, \hat{\mathbf{q}}^A_{0}, \hat{\mathbf{q}}^A_{1}, \alpha)$ }
    
    \State{\Return $\hat{\mathbf{q}}_{0}, \hat{\mathbf{q}}_{1}$}
    \EndFunction
    \end{algorithmic}
    \caption{Calibrate the decision-making thresholds}
    \label{alg:calibrate}
\end{algorithm}

The fact that different calibration samples stop at different acquisition times does not break exchangeability. Conditional on the preliminary thresholds and the acquisition policy, the stopping time is a deterministic function of the observed wafer trajectory. Hence, each calibration score is obtained by applying the same stopping time scoring rule to an exchangeable sample. The resulting stopping time scores are therefore exchangeable, even though their realized stopping times may differ. We note, however, that the resulting coverage guarantee applies directly to predictions evaluated under the stopping rule induced by the preliminary thresholds. When the refined thresholds are subsequently used to modify the stopping rule at inference, the standard split conformal exchangeability argument no longer provides an exact finite-sample coverage guarantee for the resulting adaptive procedure.

\begin{rmk}\label{rmk:alpha_inc}
    In our setting, the conformal thresholds $\hat{\mathbf{q}}_0$ and $\hat{\mathbf{q}}_1$ are used as certification thresholds. Therefore, from the definition of non-conformity scores in \Cref{eq:nonconf_scores} an increase of $\alpha$ leads to a decrease of $\hat{q}_{y, j}, \ y\in\{0,1\}, \ j=1, 2, \ldots, C$. As a result, it is harder to certify labels, and more labels remain undecided. Hence, the model requires more patches for scanning. This means that smaller thresholds lead to fewer early certifications, and therefore stopping happens later.
\end{rmk}

\begin{rmk}\label{rmk:fullscan_hybrid}
    In some cases, the condition provided by the conformal prediction framework does not hold for any $t \in \{1, 2, \ldots, T\}$. This results in the acquisition process continuing until all measurements (patches) have been obtained. If this occurs for a given wafer map, we can utilize a CNN model to predict the defect type. Since all patches are already available, this does not increase the number of scanning steps but can significantly improve inspection performance. A system that works in this manner is referred to as a \textit{hybrid} model (a similar approach has been used in \cite{dekhovich2025supconwi}). 
\end{rmk}

\section{Experiments}\label{sec4:experiments}

\subsection{Experimental setup}

\begin{figure}
    \centering
    \includegraphics[width=0.75\textwidth]{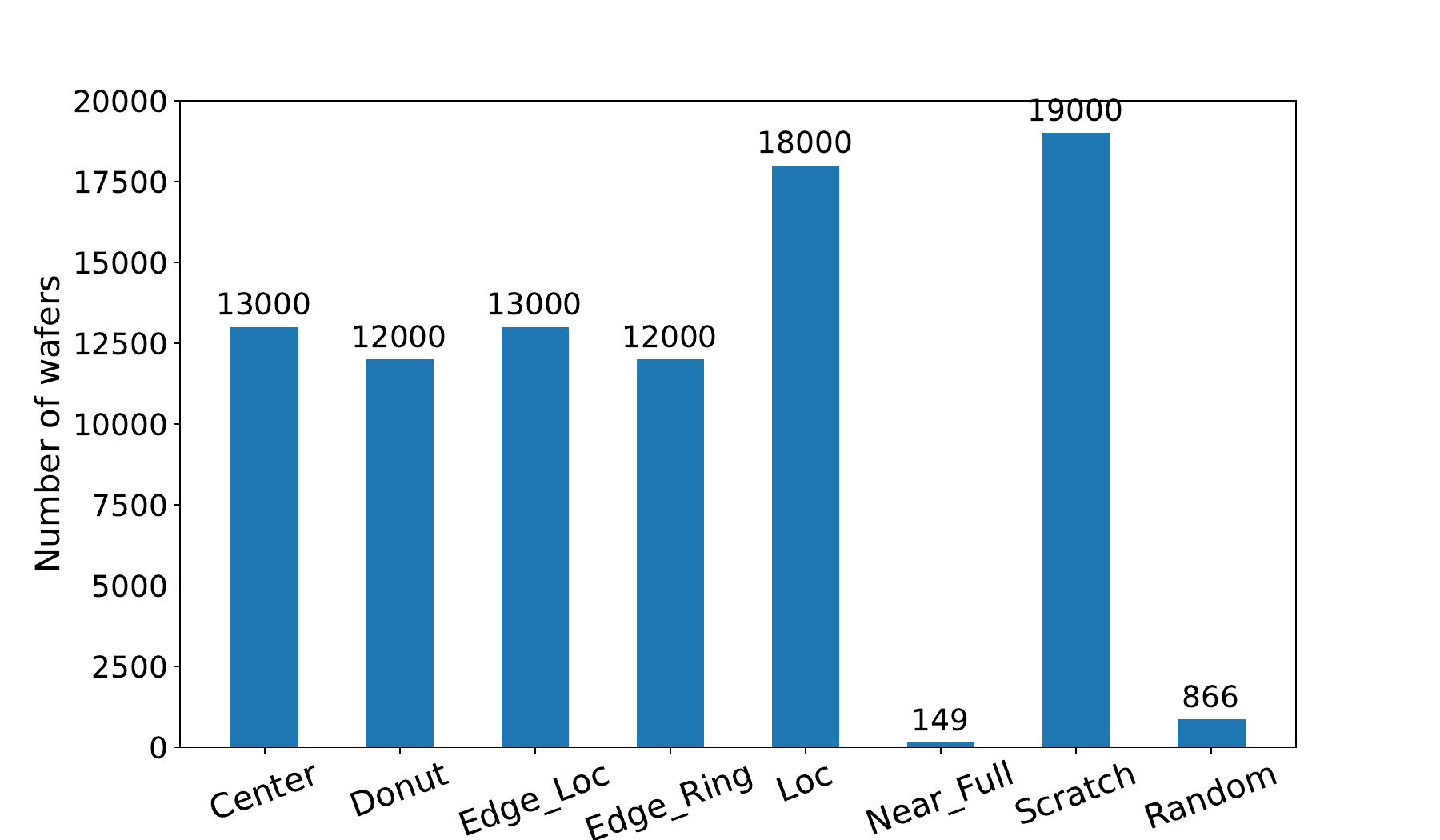}
    \caption{Number of defects of each type present in the MixedWM38 dataset.}
    \label{fig:defects_distr}
\end{figure}

\paragraph{Dataset} We validated our method on the MixedWM38 dataset, which contains around $38{,}000$ images of wafer maps with one, two, three and four defects, as well as normal ones. Each input image in the MixedWM38 dataset has a size of $52 \times 52$, which we divide into $T=169$ patches of size $4 \times 4$. Before that, we apply two sets of augmentation transformations, $\mathcal{T}_1$ and $\mathcal{T}_2$, which are required for the contrastive loss computation. These augmentations include horizontal and vertical flips with probability $0.5$, random rotations in the range $[-90, 90]$ degrees, and normalization. \Cref{fig:defects_distr} shows how many times each defect type occurs in the dataset, illustrating class imbalance and highlighting the need for robust evaluation metrics such as the F1 score.

\paragraph{Models}

Overall, we utilize two deep learning models that are trained sequentially: first, we pretrain the classifier---a causal Transformer---on random patch sequences; second, we train an RL-based patch selector that optimizes the acquisition process given the pretrained classifier; finally, we learn the conformal prediction thresholds. \Cref{tab:exp_setup} summarizes how the dataset was split to train each of the components.

\begin{table*}[ht]
    \centering
    \caption{Number of wafer maps in each subset that are used for training and evaluation.}
    \begin{tabular}{c c c}
    \toprule
     Dataset & Purpose & \# of images\\
    \midrule
        $\mathcal{D}_1$  & train transformer & $20{,}528$\\
        $\mathcal{D}_2$  & learn conformal prediction thresholds $\hat{\mathbf{q}}_0$, $\hat{\mathbf{q}}_1$  &  $2,281$\\
        $\mathcal{D}_3$  & train patch selector &  $7{,}603$\\
        $\mathcal{D}_4$  & test both transformer and patch selector & $7{,}603$\\
    \bottomrule
    \end{tabular}
    \label{tab:exp_setup}
\end{table*}

For the Transformer architecture, we employ three different configurations, which we refer to as \textit{small}, \textit{medium}, and \textit{large} based on the number of learnable parameters. \Cref{tab:transformer_diff} highlights the differences among these three configurations. Here, \texttt{embedding\_dim} denotes the dimension of $\mathbf{z}_t$, and $L$ represents the number of Transformer blocks (see \Cref{fig:ufld-transformer}). Each of the configurations is pretrained for $240$ epochs using a learning rate of $0.001$. Additionally, we employ KV-caching to speed up the inference stage. More detailed information is available in our code implementation.

\begin{wraptable}{r}{6.5cm}
    \centering
    \caption{Difference among the three Transformer configurations: small, medium and large.}
    \begin{tabular}{c c c}
    \toprule
     Capacity & embedding\_dim & $L$\\
    \midrule
        small & 128 & 8 \\
        medium & 256 & 4 \\
        large & 256 & 8\\
    \bottomrule
    \end{tabular}
    \label{tab:transformer_diff}
\end{wraptable}

To train the RL agent, we utilize the PPO algorithm \cite{schulman2017proximal}, where both the actor and critic networks are implemented as two-layer MLPs. Additionally, we employ a memory buffer to store transitions and stabilize the learning process. The agent is optimized for $100$ iterations using the Adam optimizer with a learning rate of $0.001$, the penalty term $c_{\text{RL}}=0.005$.

As baseline methods, we use SupConWI-RL \cite{dekhovich2025supconwi} and its variants. Additionally, we compare our results with the ResNet-34 model \cite{he2016deep}, a standard convolutional neural network (CNN) baseline. It is worth noting that ResNet-34 requires the full image to be acquired before processing, which makes it less efficient in terms of scanning time. However, it achieves state-of-the-art accuracy on this dataset and serves as an upper-bound benchmark.

\paragraph{Evaluation metrics} To validate our model and compare it with others, we focus on $\text{Accuracy}=\frac{ \# \{i=1, 2, \ldots, n_{\text{test}}: \ \mathbf{y}_i = \hat{\mathbf{y}}_i \}}{n_{\text{test}}}$ (therefore, an image is correctly classified only if all components of the output $\hat{\mathbf{y}} \in \{0, 1\}^{C}$ are correctly predicted), $\text{F1} = \frac{2 \cdot \text{Recall} \cdot \text{Precision}}{\text{Recall} + \text{Precision}}$, where $\text{Recall} = \frac{\text{TP}}{\text{TP} + \text{FN}}$ and $\text{Precision} = \frac{\text{TP}}{\text{TP} + \text{FP}}$ to measure how well the model predicts despite class imbalance. Here, \text{TP}, \text{FP}, and \text{FN} are the number of true positives, false positives, and false negatives, respectively, and $t^{*}$ is the number of scanning steps.

Our objective is to evaluate the trade-off between performance (accuracy and F1 score) and the scanning time ($t^{*}$) required by different models. CNNs typically achieve state-of-the-art results by utilizing the full wafer surface. Therefore, our goal is to identify a model that achieves comparable or better accuracy/F1 score while significantly reducing the required scanning time.

\subsection{Results}

First, we compare the LSTM, the proposed Transformer, and CNN models when all information is available, i.e., all patches have been acquired. \Cref{tab:baselines} summarizes the test accuracy on the MixedWM38 dataset when a full sequence of measurements is given to the classifier. The Transformer-based architectures achieve higher classification accuracy than the LSTM. It is noteworthy that all three transformer configurations give similar results despite significant differences in the number of parameters. Meanwhile, the CNN outperforms the LSTM and Transformer-based models. Nevertheless, the latter allow for predictions based on partial observations, as we show below. Additionally, the CNN model has significantly more parameters to train and requires more memory to store.

\begin{table}[ht]
    \centering
    \caption{Comparison of models in terms of accuracy and number of trainable parameters if all measurements are available to the network. For the Transformer, three different configurations are considered. The results are averaged over five independent runs, providing mean and standard deviation across runs.}
    \begin{tabular}{l c c}
    \toprule
     Model  &  Accuracy (\%) & \# of learnable parameters\\
    \midrule
        LSTM & $96.35 \pm 0.25$ & $1{,}508{,}232$ \\
    \midrule
        \multirow{3}{*}{Transformer} 
            & $98.02 \pm 0.14$ & $1{,}641{,}160$\\
            & $98.16 \pm 0.06$ & $3{,}285{,}384$\\
            & $98.10 \pm 0.11$ & $6{,}444{,}424$ \\
    \midrule
        CNN & $98.79 \pm 0.09$ & $21{,}348{,}168$ \\
    \bottomrule
    \end{tabular}
    \label{tab:baselines}
\end{table}

\Cref{tab:result} shows the results when applying our adaptive acquisition strategy compared to the SupConWI-RL variants. Our UFLD achieves significantly higher accuracy while maintaining a comparable scanning rate. Moreover, UFLD requires two deep learning models to be trained---the classifier and the RL agent---which is a more parameter-efficient strategy compared to SupConWI-RL, where the confidence estimator is another black-box deep learning model. Instead, conformal prediction provides a simple, fast, and justified approach as a stopping rule. An additional analysis of the effect of the miscoverage rate $\alpha$ is presented in \Cref{sec:cp_hyperparams}.

\begin{table*}[ht]
    \footnotesize
    \centering
    \caption{Comparison of the proposed UFLD with other methods. For single-model cases, the results are averaged over five independent runs, providing mean and standard deviation across runs.}
    \begin{tabular}{l l c c c}
    \toprule
     & Model  &  Accuracy (\%) & Scanned patches (\%) & \# of parameters\\
    \midrule
    CNN & ResNet-34 & $98.79 \pm 0.09$ & $100$ & $21{,}348{,}168$\\
    \midrule
    \multirow{4}{*}{Non-hybrid} 
        & Single SupConWI-RL & $95.58 \pm 0.26$ & $59.3 \pm 2.6$ & $3{,}194{,}554$\\
        & Ensemble SupConWI-RL & $96.34$ & $60.4$ & $15{,}972{,}770$\\
    \cline{2-5}
        & Single UFLD-large (ours) & $97.88 \pm 0.07$ & $61.4 \pm 1.6$ & $7{,}174{,}194$ \\
        & Ensemble UFLD-small (ours) & $97.82$ & $59.7$ &  $11{,}342{,}650$\\
    \midrule
    \multirow{4}{*}{Hybrid} 
        & Single SupConWI-RL & $97.74 \pm 0.20$ & $59.3 \pm 2.6$ & $24{,}542{,}722$\\
        & Ensemble SupConWI-RL & $98.45$ & $60.4$ & $37{,}320{,}938$\\
    \cline{2-5}
        & Single UFLD-large (ours) & $98.38 \pm 0.10$ & $61.4 \pm 1.6$ &  $28{,}522{,}362$\\
        & Ensemble UFLD-small (ours) & $98.37$ & $59.7$ & $32{,}690{,}818$ \\
    \bottomrule
    \end{tabular}
    \label{tab:result}
\end{table*}

First, we compare results in the non-hybrid setting (see Remark \ref{rmk:fullscan_hybrid}), i.e., the case where we do not use CNN even if the whole surface was scanned. UFLD outperforms SupConWI-RL in both single-model and ensemble cases. Moreover, even single UFLD achieves $1.5\,\mathrm{p.p.}$ higher accuracy than Ensemble SupConWI-RL, while having half as many trainable parameters.

In the hybrid setting, we do not observe much difference in either accuracy or scanned patches between UFLD variants and Ensemble SupConWI-RL, while the latter requires more trainable parameters. Single SupConWI-RL loses to our UFLD by $0.64\,\mathrm{p.p.}$ without significant advantage in scanning time. However, UFLD provides a trade-off between scanning time and accuracy that is controlled by $\alpha$, which we analyze in \Cref{sec:cp_hyperparams}. This flexibility allows for an adjustment of the inspection policy based on the specific sensitivity requirements of wafer production.

\begin{figure}
\centering
  \includegraphics[width=\linewidth]{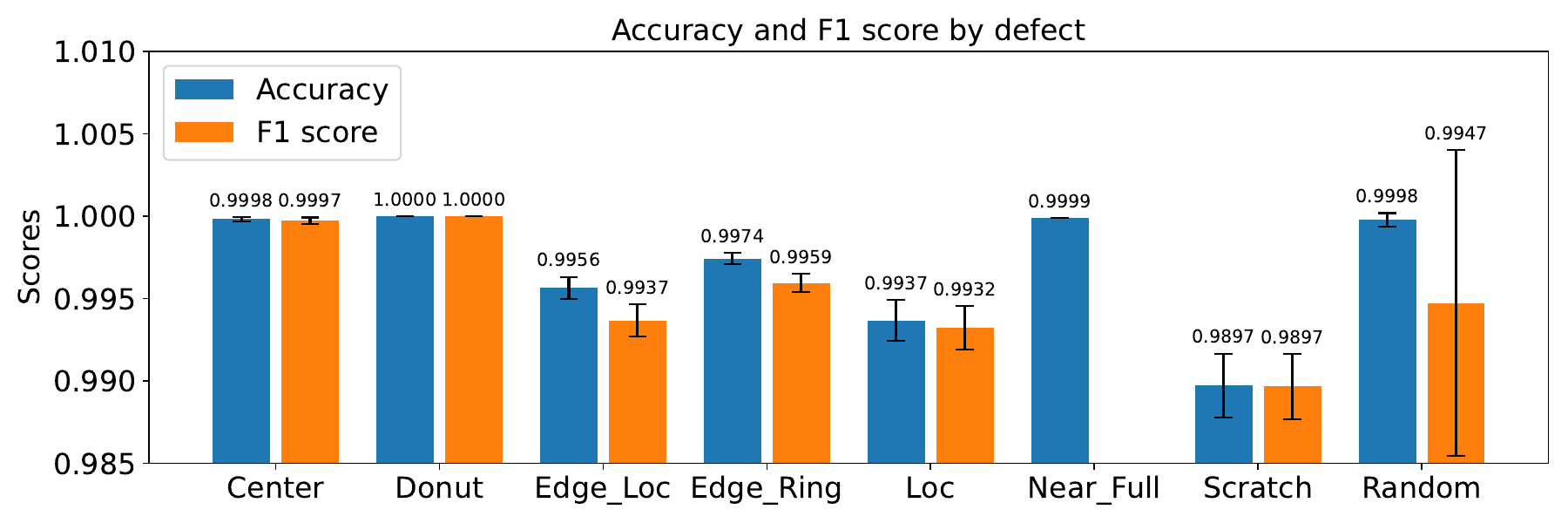}
  \caption{Accuracy and F1 score of UFLD-large calculated by defect and averaged over five runs with 2-sigma error bars.}
  \label{fig:acc_f1_by_defect}
\end{figure}

To analyze our results further, we demonstrate that UFLD effectively handles the class imbalance present in the MixedWM38 dataset. \Cref{fig:acc_f1_by_defect} shows the accuracy and F1 score for predicting each defect separately. We can see consistency in both metric values across all defect types, which illustrates the robustness of UFLD despite class imbalance. Notably, we observe a slightly lower F1 score for \textit{Near\_Full}, which can be explained by the very small number of these examples in the dataset.

\begin{figure}[th]
    \centering
    \includegraphics[width=0.8\linewidth]{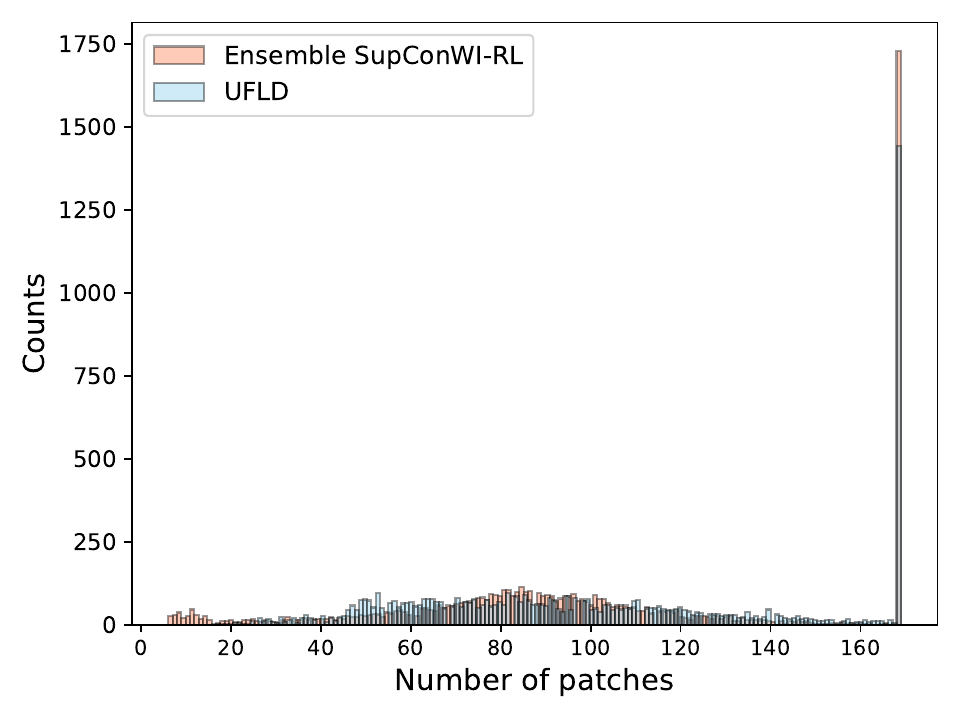}
    \caption{Comparison between Ensemble SupConWI-RL and UFLD (large) in the number of scanned patches.}
   \label{fig:measurements_comparison}
    \label{fig:placeholder}
\end{figure}

In addition, we compare the overall number of scanned patches between our UFLD (large) and the Ensemble SupConWI-RL in \Cref{fig:measurements_comparison}. Despite the equal average percentages of scanned surface (i.e., the number of patches), the behavior is different---UFLD has significantly fewer cases where it cannot stop the scanning process early. Ensemble SupConWI-RL more frequently reaches the maximum budget of patches due to higher uncertainty than UFLD. This hints that the conformal prediction framework provides a more robust and better-calibrated confidence threshold than the auxiliary confidence estimator counterpart. 

\subsection{Sensitivity of conformal prediction to the choice of hyperparameters}\label{sec:cp_hyperparams}

We want to show that our approach is robust to the choice of $\alpha$, regardless of transformer architecture. \Cref{tab:cp_alpha} demonstrates how coverage, percentage of scanned regions, and two types of accuracy change with small changes in the miscoverage level $\alpha$ for three variants of UFLD -- small, medium and large. The results show a monotonic increase in accuracy and a decrease in coverage with the increase of $\alpha$; the number of scanned patches also increases with the increase of $\alpha$. This is consistent with our two-rule conformal stopping criterion: increasing $\alpha$ decreases the conformal thresholds $\hat{\mathbf{q}}_0$ and $\hat{\mathbf{q}}_1$, making positive or negative label certification stricter (see Remark \ref{rmk:alpha_inc}). As a result, fewer labels are certified at early acquisition steps, and the RL policy must inspect more regions before meeting the stopping condition. This longer acquisition process provides the classifier with more information, which leads to higher accuracy. We observe the same behavior across all three UFLD variants, meaning that the stopping mechanism is stable with respect to model architecture. Here we report the empirical micro-averaged coverage at stopping time:

$$
\text{Coverage} = \frac{1}{NC}\sum_{i=1}^N \sum_{j=1}^C \mathbf{1}\{S_{i, \tau_{i}, j} \le \hat{q}_{y^i_j, j} \},
$$
where $\tau_i$ is the stopping time of wafer $i$.

\begin{table*}[th]
    \tiny
    \centering
    \caption{Results of UFLD models of different sizes (1.6M, 3.3M, and 6.4M), varying $\alpha$ from $0.05$ to $0.1$ with a step size of $ 0.005$.}
    \begin{tabular}{l l p{0.2in} p{0.2in} p{0.2in} p{0.2in} p{0.2in} p{0.2in} p{0.2in} p{0.2in} p{0.2in} p{0.2in} p{0.2in}}
    \toprule
        UFLD &  &  $0.05$ & $0.055$ & $0.06$ &  $0.065$ &  $0.07$ & $0.075$ & $0.08$ &  $0.085$ & $0.09$ & $0.095$ & $0.1$\\
    \midrule
    \multirow{4}{*}{small} 
        & Non-hybrid Coverage  (\%) &  $97.44$ & $97.13$ & $96.74$ & $96.36$ & $95.94$ & $95.50$ & $95.03$ & $94.54$ & $94.04$ & $93.60$ & $93.04$ \\
        & Scanned (\%) &  $54.62$ & $56.57$ & $58.54$ & $60.21$ & $61.75$ & $63.23$ & $64.51$ & $66.00$ & $67.43$ & $68.55$ & $69.69$ \\
        & Accuracy (\%) &  $97.34$ & $97.44$ & $97.51$ & $97.56$ & $97.59$ & $97.64$ & $97.67$ & $97.70$ & $97.71$ & $97.72$ & $97.74$ \\
        & Hybrid Accuracy (\%) & $97.90$ & $98.06$ & $98.18$ & $98.26$ & $98.31$ & $98.38$ & $98.42$ & $98.46$ & $98.50$ & $98.51$ & $98.55$ \\
    \midrule
    \multirow{4}{*}{medium} 
        & Non-hybrid Coverage (\%) &  $97.18$ & $96.79$ & $96.48$ & $96.13$ & $95.72$ & $95.30$ & $94.87$ & $94.50$ & $94.04$ & $93.49$ & $93.01$\\
        & Scanned (\%) &  $55.45$ & $57.30$ & $58.68$ & $60.21$ & $61.76$ &  $63.25$ & $64.54$  & $65.72$ & $67.08$ & $68.42$ & $69.58$ \\
        & Accuracy (\%) & $97.58$ & $97.68$ & $97.72$ & $97.73$ & $97.78$ &  $97.82$ & $97.83$ & $97.85$ & $97.87$ & $97.87$ & $97.88$ \\
        & Hybrid Accuracy (\%) &  $98.01$ & $98.13$ & $98.20$ & $98.22$ & $98.28$ & $98.33$ & $98.37$ & $98.40$ &  $98.44$ & $98.46$ &  $98.47$\\
    \midrule
    \multirow{4}{*}{large} 
        & Non-hybrid Coverage (\%) & $97.34$ & $96.97$ & $96.60$ & $96.25$ & $95.85$ & $95.47$ & $95.05$ & $94.60$ & $94.25$ & $93.83$ & $93.34$\\
        & Scanned (\%) & $56.3$ & $58.3$ & $60.0$ & $61.4$ & $62.9$ & $64.2$ & $65.5$ & $66.9$ & $68.0$ & $69.0$ & $70.1$\\
        & Accuracy (\%) & $97.72$ & $97.80$ & $97.85$ & $97.88$ & $97.91$ & $97.94$ &  $97.94$ & $97.96$ & $97.99$ & $98.01$ & $98.03$ \\
        & Hybrid Accuracy (\%) & $98.18$ & $98.29$ & $98.34$ & $98.38$ & $98.44$ & $98.48$ & $98.50$ & $98.52$ & $98.55$ & $98.58$ & $98.60$\\
    \bottomrule
    \end{tabular}
    \label{tab:cp_alpha}
\end{table*}

\subsection{Ablation study}

In addition to the comparison with the baselines, we conduct an ablation study to validate the contribution of each of the components in UFLD. \Cref{tab:ablation study} summarizes the classification accuracy and F1 score of a Causal Transformer classifier under different configurations. First, we evaluate the classifier that receives $61.5\%$ of the patches using random sequences and RL-based adaptive data acquisition (ADA). This percentage matches the amount of surface that UFLD needs on average for reliable predictions. We perform this comparison to validate the effect of ADA. Second, we consider the configuration where the classifier receives the random sequence of patches but uses the conformal prediction (CP) stopping rule for certainty in the predictions. The third configuration corresponds to the UFLD algorithm, where we utilize both of these techniques.

Overall, we observe a significant performance improvement when we add each of the components: Causal Transformer alone performs the worst with the given data budget, while adding RL-based ADA increases both accuracy ($+4.19 \,\mathrm{p.p.}$) and F1 score ($+0.008$), maintaining the same data budget of $61.5\%$. The classifier with the CP-based stopping rule without input sequence optimization also improves the indicators ($+4.49 \,\mathrm{p.p.}$ for the accuracy and $+0.007$ for the F1 score) but leads to an increase of $9 \,\mathrm{p.p.}$ in required patches. Finally, the employment of both ADA and the CP-based stopping rule boosts both accuracy ($+5.08\,\mathrm{p.p.}$) and the F1 score ($+0.008$) utilizing the same percentage of the wafer surface ($61.4\%$).  

\begin{table}[ht]
    \centering
    \small
    \caption{Ablation study results where the Transformer classifier (large) with or without adaptive data acquisition (ADA) and conformal prediction (CP) stopping rule are considered. The results are averaged over five independent runs.}
    \begin{tabular}{l c c c c c}
    \toprule
     Model & RL  &  CP & Acc (\%) & F1 & Scanned (\%)\\
    \midrule
      Transformer & no & no & $92.80$ & $0.985$ & $61.5$\\ 
       + ADA & yes & no & $96.99$ & $0.993$ & $61.5$\\
       + Stopping rule & no & yes & $97.29$ & $0.992$ & $70.5$ \\
    \midrule
     UFLD & yes & yes & $97.88$ & $0.993$ & $61.4$\\ 
    \bottomrule
    \end{tabular}
    \label{tab:ablation study}
\end{table}

\section{Conclusion}\label{sec5:conclusion}

In this work, we addressed the wafer map defect classification problem using an adaptive data acquisition approach. We propose UFLD, a reinforcement learning-based algorithm that adaptively selects the next location for the measurement. In order to make the acquisition process more efficient and reduce operational costs, we employ a conformal prediction-based stopping rule. In contrast to previous work in the semiconductor domain, which requires an additional black-box deep learning model for confidence estimation, the conformal prediction framework is more parameter-efficient. 
In addition, the proposed UFLD uses a causal transformer architecture as a classifier, instead of RNN-based architectures that were utilized in previous works. This results in a $1.5\,\mathrm{p.p.}$ increase in diagnostic accuracy, requiring around $60\%$ of the surface, which is the same number of measurements as previous methods. 
Moreover, single UFLD uses $50\%$ fewer parameters compared to the baselines due to the elimination of the auxiliary confidence estimator network. Therefore, it further closes the gap to CNN-based approaches that require full scanning. Additionally, variations in the miscoverage rate provide a controllable trade-off between inspection performance and scanning time.

In our opinion, the UFLD framework contributes to more reliable and cost-efficient semiconductor manufacturing. By reducing the need for full surface scanning, our approach increases the throughput of inspection tools in high-volume manufacturing settings. Furthermore, conformal prediction provides a principled mechanism for constructing prediction sets per label and defining the stopping criterion through an interpretable miscoverage parameter $\alpha$, compared to black-box deep learning methods. Future work could improve the current framework by developing more rigorous theoretical guarantees and by exploring extensions of this adaptive acquisition strategy to other problems, such as defect segmentation or surface reconstruction.

\begin{ack}
The project is supported by the Chips Joint Undertaking and its members, including the top-up funding by RVO (The Netherlands Enterprise Agency).
\end{ack}


\bibliographystyle{unsrt}  
\bibliography{references.bib}



\end{document}